\documentclass{article}

\usepackage{verbatim,soul}

 \usepackage[nonatbib,preprint]{nips_2018}

\usepackage[utf8]{inputenc} 
\usepackage[T1]{fontenc}    
\usepackage{hyperref}       
\usepackage{url}            
\usepackage{booktabs}       
\usepackage{amsfonts}       
\usepackage{nicefrac}       
\usepackage{microtype}      
\usepackage{wrapfig}

\usepackage[cmex10]{amsmath}
\usepackage{amsthm}
\DeclareMathOperator*{\argmax}{argmax}

\usepackage{paralist,xparse}
\usepackage{dsfont}
\usepackage{amsfonts}
\usepackage{multirow} 
\usepackage{subfigure}
\usepackage{amssymb} 
\usepackage{scrextend}
\usepackage{float}

\usepackage{algorithm}
\usepackage{algorithmic}
\usepackage{mathtools}
\usepackage{times}
\usepackage{pgfplots}
\usepackage{tikz}
\usepackage[european resistor, european voltage, european current]{circuitikz}
\usetikzlibrary{arrows,shapes,positioning,snakes}
\usetikzlibrary{decorations.markings,decorations.pathmorphing,decorations.pathreplacing}
\usetikzlibrary{calc,patterns,shapes.geometric,topaths}

\newtheorem{theorem}{Theorem}
\newtheorem{definition}{Definition}

\newcommand{\E}{\Bbb{E}}

\title{Estimating Rationally Inattentive Utility Functions with Deep Clustering for Framing - Applications in YouTube Engagement Dynamics}

\author{William~Hoiles,\\
  \texttt{whoiles@ece.ubc.ca} \\
 \And
Vikram Krishnamurthy \\
 Cornell University
\texttt{vikramk@cornell.edu}}

\begin{document}

\maketitle

\begin{abstract} We consider a  framework involving behavioral economics  and machine learning.  Rationally inattentive Bayesian agents     make decisions based on their posterior distribution, utility function and information acquisition cost (R{\'e}nyi divergence which generalizes Shannon mutual  information). By observing these decisions, how can an observer estimate the utility function and information acquisition cost?  
 Using deep learning, we estimate framing information (essential extrinsic  features) that determines  the agent's attention strategy. Then
  we present a preference based  inverse reinforcement learning algorithm to test for rational inattention: is the agent an utility maximizer, attention maximizer, and does an information cost function exist that  rationalizes the data? The test  imposes a R{\'e}nyi mutual information  constraint which  impacts how the agent can select attention strategies to maximize their expected utility.   The test   provides  constructive estimates of the utility function and information acquisition  cost  of the agent. We illustrate  these methods on  a massive YouTube dataset for characterizing the commenting behavior of users.
\end{abstract}

\section{Introduction}
\label{sec:Introduction}

Suppose  a Bayesian agent  chooses an  action at each time instant to maximize an expected utility function based on the noisy measurement of an underlying state. Assume that obtaining this noisy measurement is expensive  -- 
this information acquisition  cost affects the action chosen by the agent.  An observer records the dataset of actions  of the Bayesian agent and knows the underlying state. How can the observer estimate the utility function  and information acquisition cost of the agent given this dataset?
Our aim is to construct preference based inverse reinforcement learning
algorithms to obtain set valued estimates of the utility and information acquisition cost that are consistent with the  dataset. 

Our methodology  stems from behavioural economics and machine learning:  non-parametric estimation   of utility  functions
and feature extraction using deep clustering to construct
behavioural-economics based models  for  Bayesian agents. Let us briefly explain these two aspects.
Estimating  utility functions given a finite length time series of decisions is well studied in the area of revealed preferences  in economics~\cite{Var12,Woo12} and more recently  in machine learning. Also,
costly  information acquisition by Bayesian agents has been 
 studied by  economists and psychologists under the area of ``rational inattention'' pioneered by Sims~\cite{Sim03,Sim10}. Rational inattention is a form of bounded rationality - the key idea is that human attention spans for information acquisition are limited and  can be modelled in information theoretic terms as a Shannon capacity limited communication channel. However, modelling  the information acquisition process is complicated in our case by framing. In behavioural economics, Kahneman uses  ``frames''  to describe information  an agent has when making a decision.  For example, when selecting which product to purchase on a website, the positioning of the products and surrounding content on the website  impacts how humans select a product. Given external information (image/text/numeric) in which the decision problem is embedded, how can one construct a tractable feature set?
 We develop deep embedded clustering methods to  construct the  frames to  test for rational inattentive agents. The deep embedded clustering  is based on~\cite{XGF16,GGLY17}, however we  design the input, encoder, and decoder to account for the visual perception of the frame of the decision problem which includes image, text, and numeric information.

{\em Context: (i) Rational Inattention \& Inverse Reinforcement Learning}. 
Sim's rational inattention model is studied extensively in behavioral economics~\cite{MM15}. Woodford~\cite{Woo12} considered an upper bound on the Shannon capacity  for testing  rational inattention with visual perception queues. Typically, the information acquisition  costs faced by a decision maker are not known to the observer. A general test for rational inattention is proposed in~\cite{CM15,CD15} with minimal restrictions on the  information acquisition  cost.
The two significant extensions considered in this paper are the effects of framing
(determined using deep embedded clustering) and the use of R{\'e}nyi mutual information cost constraints for testing rational inattention.
Our rational inattention test  is equivalent to solving the temporal credit assignment problem  in {\em preference-based inverse reinforcement learning}~\cite{WANF17}. Such inverse reinforcement learning
is used with non-numeric feedback~\cite{WFN16}, e.g.\ in socially adaptive path planning~\cite{KP16,HMARD17} for robots.

{\em Context: (ii) YouTube Application}.
We will use rational inattention and framing
(with deep learning) on a massive YouTube data set  to analyse  the commenting behaviour of users in YouTube.
 Extensive  studies  \cite{Kha17,HG17,ABCH15} show  that comments 
 posted by users
are  influenced by the thumbnail, title, category, and perceived popularity of each video.
 In our formulation,  frames are associated with the videos thumbnail and title; the decision-problem with the category; and the perceived popularity with the underlying state. The  commenting behavior (agent's actions) is related to the number and sentiment of the comments that result from the framing information, state, and decision-problem faced by the agent. Based on  extensive
data analysis,  our main take-home message (from a behavioral economics point of view)  is  that YouTube users are rationally inattentive in their  commenting behavior; moreover users prefer to comment on videos that are perceived to be popular; see Sec.\ref{sec:RationalInattentionandUtilityMaximizationintheYouTubeSocialNetwork} for additional conclusions.



{\em Organization}.  Sec.\ref{sec:ProblemFormulationandRationalInattention} introduces the problem formulation.
Sec.\ref{sec:FrameingInformationDeepLearning} discusses a deep embedded clustering algorithm for associating the observed agent's action  to specific frames. 
In Sec.\ref{sec:DecisionTestforRationalInattentionandRecoverabilityofAgentPreferences} and  \ref{sec:MutualInformationMeasuresandAttentionCost}, the  decision test for rational inattention with R{\'e}nyi mutual information acquisition cost  are provided. The tests are constructive: they provide estimates
of the  utility function, information acquisition  cost, and attention strategy.
Sec .\ref{sec:Risk-AwareUtilityMaximizationbyOff-PolicyEstimation} provides  Bernstein based  finite sample performance bounds.
Sec.\ref{sec:RationalInattentionandUtilityMaximizationintheYouTubeSocialNetwork} applies the methods  to a massive  YouTube dataset to  characterize the commenting behavior of users.
The appendix summarizes the implementation details of the deep classifier.

\section{Problem Formulation and Rational Inattention}
\label{sec:ProblemFormulationandRationalInattention}
We first describe the problem formulation first from the  point of view of the  rationally inattentive agent; and then from the point of view of the observer that views the dataset generated by the agent. Despite our abstract formulation, the reader should keep in mind  the YouTube context outlined above.

\subsubsection*{Viewpoint 1. Rationally Inattentive Bayesian Agent}
Assume the agent knows the finite state space  $\mathcal{X}$ and finite action space $\mathcal{A}$. The agent's prior beliefs of the possible states are given by the prior probability distribution $\mu(x)$, $ x\in \mathcal{X}$.
The {\em  attention function} $\alpha(s|x)$ of  the agent  provides a distribution over the signals $s\in\mathcal{S}(\alpha)$ when the state is $x$. The set of possible signals $\mathcal{S}(\alpha)$ for a given attention strategy $\alpha$ is finite. The attention function encodes all the information (signals, private information, and measurement mechanism) available to the agent to compute the   posterior state distribution. Given the prior  $\mu(x)$, and attention function $\alpha(s|x)$, the Bayesian agent computes the posterior distribution as
\begin{equation}
p(x|s) = \frac{\mu(x)\alpha(s|x)}{\sum\limits_{y\in\mathcal{X}}\mu(y)\alpha(s|y)}.
\label{eqn:stateposterior}
\end{equation}
The  agent has utility function $u(x,a)$ over the states $x\in\mathcal{X}$ and actions $a\in\mathcal{A}$.
\begin{definition}
  An agent satisfies attention rationality if it selects actions $a\in\mathcal{A}$ and attention functions $\alpha(s|x)$ that satisfy the following conditions
  (where $\E$ denotes the expectation operator):
\begin{compactenum}[i)]
\item Expected Utility Maximization:
\begin{equation}
a^* \in\argmax_{a \in \mathcal{A}} \E\{ u(x,a) | s\} = \argmax_{a \in \mathcal{A}} \left\{\sum\limits_{x\in\mathcal{X}}p(x|s)u(x,a)\right\} \quad \forall p(x|s)\in\mathcal{S}(\alpha)
\label{eqn:utilitymaximization}
\end{equation}
\item Attention Selection Rationality: 
\begin{align}
&\alpha^*(s|x) \in\argmax_{\alpha} \Big\{\E_{s\in\mathcal{S}(\alpha)}\{\operatorname*{max}_{a\in\mathcal{A}}[\sum_{x\in\mathcal{X}}p(x|s)u(x,a)]\}-
 C(\mu,\alpha)\Big\}
\label{eqn:attentionmaximization}
\end{align}
where $C(\mu,\alpha)$ is the cost (or disutility) of attention function $\alpha$ when the prior distribution $\mu$. 
\end{compactenum}
\label{def:rationalinattention}
\end{definition}
\noindent
Eq.(\ref{eqn:utilitymaximization}) states that the agent  selects actions that are consistent with Bayesian utility maximization, and (\ref{eqn:attentionmaximization}) states that the agent selects the best attention strategy to maximize the gross expected utility. 

\subsubsection*{Viewpoint 2. Observer's Model and Deep Clustering of Frames}
By observing the actions of the agent,  the observer aims to determine
if the agent is rationally inattentive, and if so, estimate the agent's utility function and information acquisition  cost. The observer has access to the  dataset of states $x_t$ and actions $a_t$ chosen by the agent for time
$t=1,\ldots, T$:
\begin{equation}
\mathcal{D} = \{(x_t,f_t,a_t)\}_{t=1}^T.
\label{eqn:observationdataframe}
\end{equation}
Here the parameter $f_t$  represents all the framing information immediately apparent to the agent. Typically, framing information $f_t$  includes images, video, text, and  data. In our YouTube example, $f_t$ maps the title and thumbnail of a video to
an  integer representing a unique frame. Qualitatively, different values of $f_t$ determine different action policies  by the agent for a given title and thumbnail.
A major challenge when applying rational inattention theory is accounting for the agent's framing effects that impact the agent's behaviour. To account for framing effects, we assume there are $\{0,1,\dots,N\}$ possible  frames. In Sec.\ref{sec:FrameingInformationDeepLearning} a deep embedded clustering method is used to construct $f_t$ given the title and thumbnail of the YouTube video observed at time $t$. 

Given the set of frames,  rational inattention theory aims is to determine if the dataset $\mathcal{D}$ is consistent with a rational agent (Definition~\ref{def:rationalinattention}). To test for rational inattention we require  estimates of the (possibly randomized)  action selection policy $\pi(a|x,f)$ and prior beliefs $\mu(x)$ of the agent. Using $\mathcal{D}$
\begin{equation}
\hat{\pi}(a|x,f) = \frac{\sum_{t=1}^T\mathbf{1}\{x_t=x,a_t=a, f_t=f\}}{\mathbf{1}\{x_t=x, f_t = f\}}, \quad
\hat{\mu}(x) = \frac{1}{T}\sum_{t=1}^T\mathbf{1}\{x_t=x\}
\label{eqn:policyprior}
\end{equation}
are maximum likelihood estimates of these,
where $\mathbf{1}\{\cdot\}$ is the indicator function. Given the maximum likelihood estimates (\ref{eqn:policyprior}), Sec.\ref{sec:DecisionTestforRationalInattentionandRecoverabilityofAgentPreferences} provides a decision test for rational inattention. For agents that satisfy the rational inattention test, methods to recover their utility function $u(x,a,f)$, attention strategy $\alpha(s|x)$, posterior distribution $s(x)$, and information cost $C(\mu,\alpha)$ are provided. 

For a rationally inattentive agent, it is desirable to have a risk-aware method to optimize the expected utility of the agent by adjusting their action selection policy $\pi(a|x,f)$ while keeping the attention strategy $\alpha(s|x)$ (measurement device) unchanged. The expected utility of a rationally inattentive agent for  action-selection policies $\boldsymbol{\pi}(a|x,f)=\{\pi_k(a|x,f)\}_{k=1}^K$ over $K$ decision problems is
\begin{equation}
V(\boldsymbol{\pi}(a|x,f)) = \sum_{k=1}^K\sum_{x\in\mathcal{X}}\sum_{a\in\mathcal{A}_k}\pi_k(a|x,f)\mu(x)u(a,x,f).
\label{eqn:totalexpectedutility}
\end{equation}
In Sec.\ref{sec:Risk-AwareUtilityMaximizationbyOff-PolicyEstimation} a penalized variance optimization method is presented  for constructing action selection policies that maximize (\ref{eqn:totalexpectedutility}). The construction  uses finite sample bounds on the total expected utility.


\section{Constructing Preference and Policy Invariant Frames via Deep Learning}
\label{sec:FrameingInformationDeepLearning}

Here a deep embedding method is provided that learns the policy invariant frames of the agent. Specifically, a mapping of $f_t$ to $n_t\in\{1,\dots,N\}$ is constructed where for each $n\in\{1,\dots,N\}$ the behavior of the agent is invariant. In the YouTube social network the framing information available to the agent is comprised of the title and thumbnail of each video. Given that agents are ordinal preference invariant to minor variations in the title and thumbnail, it is possible to map the features $f_t$ to one of $\{1,\dots,N\}$ discrete frames learned using deep embedding. 

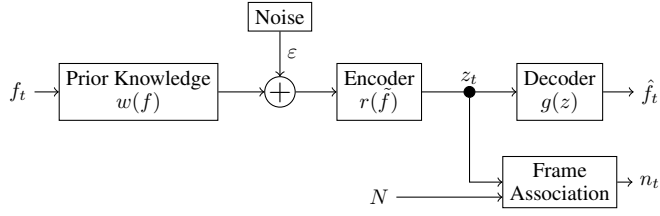
\begin{figure}[h!]
  \centering
\begin{tikzpicture}[font = \normalsize, scale =0.8,transform shape]
\tikzstyle{block} = [draw,black,rectangle,fill= white,align=center]

\def\xo{1.0}
\def\x{3.1}

\node (FI) at (-2,0) {$f_t$};
\node (FO) at (8.5,0) {$\hat{f}_t$};
\node (KO) at (8.5,-2+0.5) {$n_t$};
\node (KI) at (4,-2.25+0.5) {$N$};

\node[block] (PK) at (0,0) {Prior Knowledge \\ $w(f)$};
\node[block] (NS) at (2.35,1.25) {Noise};
\node[draw,circle,inner sep=0pt,minimum size=3pt] (N) at (2.35,0) {\Large $+$};
\node[block] (E) at (4,0) {Encoder \\ $r(\tilde{f})$};
\node[block] (D) at (7,0) {Decoder \\ $g(z)$};
\node[draw,fill=black,circle,inner sep=0pt,minimum size=2pt] (LS) at (5.5,0) {\_};
\node[block] (CL) at (7,-2+0.5) {Frame \\ Association};

\draw[black, ->] (FI) -- (PK);
\draw[black, ->] (PK) -- (N);
\draw[black, ->] (N) -- (E);
\draw[black, ->] (E) --node[midway,above]{$z_t$} (D);
\draw[black, ->] (NS) --node[midway,right]{$\varepsilon$} (N);
\draw[black, ->] (D) -- (FO);
\draw[black, ->] (LS) |- (CL);
\draw[black, ->] (CL) -- (KO);
\draw[black, ->] (KI.east) -- (KI-|CL.west);
\end{tikzpicture}
    \caption{Schematic of the deep embedded clustering method to map the framing information $f_t$ to the discrete frame $\{1,\dots,N\}$. The parameter $w(f)$ contains all prior knowledge of the input framing information, $\varepsilon$ is a Gaussian white noise term, $r(\tilde{f})$ is the encoder, $z_t$ is the latent space representation of $f_t$, $g(z)$ represents the decoder, and $\hat{f}_t$ is the output of the autoencoder.}
\label{fig:deepembeddedclustering}
\vspace{-5pt}
\end{figure}

The deep embedding method uses an autoencoder to construct the latent representation $z_t$ of $f_t$, and includes a clustering layer to simultaneously learn how to associate each $f_t$ to one of $\{1,\dots,N\}$ discrete frames. A schematic of the clustering method is illustrated in Fig.~\ref{fig:deepembeddedclustering}. 

The autoencoder comprises two deep neural networks, the first is the encoder that  maps the input $f_t$ to the latent space representation $z_t$, and the second is the decoder that map the latent space representation $z_t$ to the input $f_t$ where $\hat{f}_t\approx f_t$. To force the encoder to learn robust latent representations, the autoencoder is trained using corrupted versions of the input. Such an autoencoder is known as a denoising autoencoder~\cite{VLBM08,Ben09}. The denoising autoencoder  encodes the input into the latent space representation, and attempts to remove the effect of the corruption process stochastically applied to the input of the autoencoder. Removing effects of the corruption process is performed by learning the statistical dependencies between the inputs. A detailed description of the denoising autoencoder architecture is in the Appendix  with focus on the title and thumbnail of YouTube videos. 

Though the latent space representation of the input has been used extensively for clustering, such methods are not guaranteed to preserve any intrinsic local structure of the framing data $f_t$. To ensure the autoencoder both minimizes the reconstruction error and maximizes the intrinsic local structure of the data, a clustering loss is used. The loss of the deep embedded clustering method (Fig.~\ref{fig:deepembeddedclustering}) is:
\begin{equation}
L = ||s-g(f(w(s)+\varepsilon))||_2^2+\operatorname{KL}(P||Q)
\label{eqn:lossfunction}
\end{equation}
where $\operatorname{KL}(P||Q)$ is the Kullback-Leibler (KL) divergence  of the discrete probability distributions $P$ and $Q$. 
Here $Q$ is the prior probability distribution of cluster association between the latent variables $z_t$ and the associated frames $n_t$. If we assume each cluster is generated from a Gaussian normal distribution with mean $\Psi_n$, then the probability of association of each $z_t$ is given by the Student-t distribution:
\begin{equation}
q_{tn} = \frac{(1+||z_t-\Psi_n||^2)^{-1})}{\sum_{n=1}^N(1+||z_t-\Psi_n||^2)^{-1}} \quad \forall n\in\{1,\dots,N\}.
\label{eqn:clusterprior}
\end{equation}
Given $Q$, the distribution $P$ is designed to avoid degenerate clustering solutions which allocate most of the frames to a few clusters or assign a cluster to a sample outlier.
\begin{equation}
p_{tn} = \frac{q_{tn}^2/F_n}{\sum_{n=1}^N(q_{tn}^2/F_n)},  \quad F_n=\sum_{t=1}^Tq_{tn}.
\label{eqn:clustertargetdist}
\end{equation}
$P(z_t=n)=F_n/T$ is the probability that  $z_t$ belonging to cluster $n$; $F_n$ is the clustering frequency.  

From (\ref{eqn:clusterprior}) and (\ref{eqn:clustertargetdist}), if all the data-points are associated with a specific cluster this will increase the loss (\ref{eqn:lossfunction}). Additionally, if the cluster is associated with several data points with low-confidence, this will also increase the loss (\ref{eqn:lossfunction}). Minimizing the loss (\ref{eqn:lossfunction}) can be interpreted as a form of self-training as $P$ depends on $Q$. Specifically, in self-training we take an initial classifier and an unlabeled dataset, then label the dataset with the classifier in order to train on its own high confidence predictions. This ensures that the latent clusters are constructed to avoid outliers. 

The deep embedding method that maps $f_t$ to $n_t\in\{1,\dots,N\}$ is formalized in Algorithm~\ref{alg:framingassociation}. The pretraining step is used to initialize the encoder and decoder parameters prior to performing any clustering. This is a critical step as the initial latent space representation of $\{f_t\}_{t=1}^T$ is used to select the approximate locations of the $N$ latent space cluster centers $\Psi^o$. Given the pretrained denoising autoencoder weights, we use the Lloyd heuristic algorithm to select the locations of the $N$ latent space cluster centers $\Psi^o$. Given the cluster centers, the deep clustering method is applied to minimize the loss (\ref{eqn:lossfunction}) by simultaneously adjusting the cluster associations and autoencoder weights. Note that in Algorithm~\ref{alg:framingassociation}, since the distribution $P$ (\ref{eqn:clustertargetdist}) depends on the weights of the encoder, we update $P$ after $\zeta$ iterations. This reduces the probability of  instability associated with cycling between adjusting weights and cluster associations. The final result of Algorithm~\ref{alg:framingassociation} is achieved when the change in cluster associations is below a  threshold  $\delta$. To ensure only frames $f_t$ that can be confidently associated to one invariant frame, all frames that fail to satisfy $\operatorname{max}\{q_{tn}\} \leq \delta_c$ are discarded. 

\makeatletter
\newcommand{\HEADER}[1]{\ALC@it\underline{\textsc{#1}}\begin{ALC@g}}
\newcommand{\ENDHEADER}{\end{ALC@g}}
\makeatother

\begin{algorithm} 
\caption{Deep Embedded Clustering for Framing Association} 
\label{alg:framingassociation} 
\begin{algorithmic} 
    \REQUIRE Set of framing information $\{f_t\}_{t=1}^T$, number of unique frames $N$, stopping threshold $\delta\in(0,1)$, confidence threshold $\delta_c\in(0,1)$, and updating interval $\zeta$. 
    \HEADER{Pretrain}
    	\STATE Pretrain the denoising autoencoder without any frame association.
    \ENDHEADER
    \HEADER{Initialize}
    	\STATE Initialize the $N$ cluster centers $\Psi^o$ using k-means clustering in the latent space and set $\varepsilon=0$.
    \ENDHEADER
        \HEADER{Deep Clustering}
        \STATE Train the deep clustering autoencoder and frame association layers (refer to Supporting Material). 
    \ENDHEADER
\RETURN Invariant frames $n_t \forall t\in\{1,\dots,T\}$ such that $\operatorname{max}_n\{q_{tn}\} > \delta_c$. 
\end{algorithmic}
\end{algorithm}

Given the preference and policy invariant frames $\{n_t\}_{t=1}^T$, we substitute $n_t\rightarrow f_t$ in $\mathcal{D}$ (\ref{eqn:observationdataframe}). Using $\mathcal{D}$ with the invariant frames, Sec.\ref{sec:DecisionTestforRationalInattentionandRecoverabilityofAgentPreferences} and Sec.\ref{sec:MutualInformationMeasuresandAttentionCost} illustrate how to detect if the agent is rationally inattentive for different information cost constraints, and how to recover the utility functions.

\section{Decision Test for Rational Inattention; Estimating Utility/ Attention Costs}
\label{sec:DecisionTestforRationalInattentionandRecoverabilityofAgentPreferences}

Here we construct a decision test for rational inattention (Definition~\ref{def:rationalinattention}). The resulting preference-based inverse reinforcement learning algorithm uses  the observed stochastic choice dataset $\mathcal{D}$ (\ref{eqn:observationdataframe}) and invariant frames $\{n_t\}_{t=1}^T$. Theorem~\ref{thrm:decisiontestrationalinattention} is our main result and generalizes \cite{CD15,CM15}:

\begin{theorem}
Dataset $\mathcal{D}$ (\ref{eqn:observationdataframe}) satisfies rational inattention (Definition~\ref{def:rationalinattention}) iff  the action policy satisfies
\begin{equation}
\pi_k(a|x,f) = \sum_{s\in\mathcal{S}(\alpha_k)}\alpha_k(s|x,f)\eta_k(a|s), \quad
\mathcal{S}(\alpha_k) = \{p_k(x|a,f): a\in\mathcal{A}_k\} \nonumber
\label{eqn:datamatching}
\end{equation}
where the choice function  $\eta_k(a|s)$ is the probability of selecting action $a$ given the posterior  associated with signal $s\in\mathcal{S}(\alpha_k)$. Additionally, one of the following two conditions must be satisfied. 
\begin{compactenum}[i)]
\item The utility  $u(x,a,f)$ satisfies the following inequalities for decision
  problems $k=1,\ldots,K$:
\begin{align}
&\sum_{x\in\mathcal{X}}p_k(x|a,f)[u(x,a,f)-u(x,b,f)] \geq 0 \quad \forall a,b\in\mathcal{A}_k \quad \forall f\in\{1,\dots,N\} \nonumber\\
&p_k(x|a,f) = \frac{\mu(x)\pi_k(a|x,f)}{\sum_{y\in\mathcal{X}}\mu(y)\pi_k(a|y,f)}
\label{eqn:bayesianexpectedutility}
\end{align}
 Also, the attention function $\alpha_k(s|x,f)$ for each decision problem $k=1,\dots,K$ satisfies
\begin{align}
&\sum_{k=1}^{K}G_{k,k}-G_{k+1,k} \geq 0 \\
&G_{k,w} = \sum_{s\in\mathcal{S}(\alpha_k)}\sum_{x\in\mathcal{X}}\mu(x)\alpha_k(s|x,f)\operatorname*{max}_{b\in\mathcal{A}_w}\left\{\sum_{x\in\mathcal{X}}s(x)u(x,b,f)\right\} \nonumber\\
  &\alpha_k(s|x) =\sum_{a\in\mathcal{A}_k}\pi_k(a|x,f)\mathbf{1}\{p_k(x|a,f) = s\},
\; \text{ with $\mathcal{A}_{K+1}=\mathcal{A}_1$. }
    \nonumber
\label{eqn:attentionmaximization}
\end{align}
\item A utility function $u(x,a,f)$ exists that satisfies the constraints
\begin{equation}
 \mathcal{L}( u(x,a,f) )  \quad\text{ for }  f \in \{1,2,\ldots, N\}
 \label{eqn:MILPagentutility}
 \end{equation}
 where the mixed integer linear constraint set $\mathcal{L}$ is defined  in the Supporting Material. 
\end{compactenum}
\label{thrm:decisiontestrationalinattention}
\end{theorem}

In Theorem~\ref{thrm:decisiontestrationalinattention}, (\ref{eqn:datamatching}) ensures that the attention function $\alpha_k(s|x,f)$ and action selection policy $\eta_k(a|s)$ are consistent with the observed action-selection policy $\pi_k(a|x,f)$ (\ref{eqn:policyprior}). The inequalities (\ref{eqn:bayesianexpectedutility}) ensure that the agent satisfies Bayesian expected utility maximization. Intuitively, if the expected utility of taking action $a$ is higher then action $b$, then $u(x,a,f) \geq u(x,b,f)$.  Additionally, the utility function must satisfy ``cyclical consistency'' in which ordinal relation cycles such as $u(x,a,f) \geq u(x,b,f) > u(x,c,f) > u(x,a,f)$ are not present. For readers familiar with revealed preference theory, this is analogous to the GARP conditions in Afriat's theorem~\cite{Var12,Die12} for testing utility maximization behavior. The constraints (\ref{eqn:attentionmaximization}) ensures the optimal attention function is selected by the agent for each decision problem. Qualitatively, $G_{k,w}$ gives the expected utility of using attention strategy $\alpha_k(s|x,f)$. The constraints (\ref{eqn:MILPagentutility}) in Theorem~\ref{thrm:decisiontestrationalinattention} provides a method to simultaneously test if the agent is rationally inattentive, and to recover the ordinal utility $u(x,a,f)$ of their associated preferences. The evaluation involves determining if a feasible solutions exists for a set of mixed-integer linear constraints. 

Notice that Theorem~\ref{thrm:decisiontestrationalinattention} places no restrictions on the information cost $C(\mu,\alpha)$ of using attention function $\alpha$ when the prior is $\mu$. That is, if the constraints (\ref{eqn:MILPagentutility}) are satisfied then the constraints 
\begin{equation}
G_{k,k}- C(\mu,\alpha_k) \geq G_{w,k}- C(\mu,\alpha_w) \quad \forall k,w\in\{1,\dots,K\}
\label{eqn:attentionmaxincostequality}
\end{equation}
are guaranteed to be feasible. The constraints (\ref{eqn:attentionmaxincostequality}) ensure that the selected attention function $\alpha_k$ is optimal for the associated decision problem $(\mathcal{X},\mu,\pi_k(a|x,f),\mathcal{A}_k)$.
The constraints (\ref{eqn:attentionmaxincostequality}) can be used to recover set valued estimates of  cost structure of the attention functions via a set of linear constraints, refer to the Supporting Material.


\section{R{\'{e}}nyi Entropy Information Acquisition Cost for Rational Inattention}
\label{sec:MutualInformationMeasuresandAttentionCost}
In this section we impose a specific structure to the information acquisition cost which defines the attention strategy of a rationally inattentive agent. 
Sims' pioneering work~\cite{Sim10} uses Shannon mutual information, here the more general R{\'{e}}nyi  mutual information is considered. The R{\'{e}}nyi mutual information between the prior $\mu(x)$ of the state and the selected attention strategy $\alpha_k(s|x)$ is
\begin{equation}
I_\beta(\mu,\alpha_k) =
   \begin{dcases}
     \frac{1}{\beta-1}\operatorname{ln}\left(\sum_{x\in\mathcal{X}}\sum_{a\in\mathcal{A}}\frac{p^\beta(x,a)}{\mu^{\beta-1}(x)p^{\beta-1}(a)}\right) \quad \beta\in(0,1)\cup(1,\infty)\\
     I(\mu,\alpha_k) \quad \beta = 1\\
     -\operatorname{ln}\left(\sum_{x\in\mathcal{X}}\sum_{a\in\mathcal{A}}\mu(x)p(a)\mathds{1}\{p(x,a) > 0\}\right) \quad \beta = 0
   \end{dcases}
   \label{eqn:renyimutualinformation}
\end{equation}   
where $\beta\in[0,\infty)$ is the R{\'{e}}nyi order. An important feature of (\ref{eqn:renyimutualinformation}) is that for $\beta\in[0,1]$ the information constraint is convex in the arguments $p(x,a)$ and $\mu(x)p(a)$, and for $\beta > 1$ the information constraint is convex in $\mu(x)p(a)$ and quasi-convex in $p(x,a)$~\cite{VH14,HV15,XE10}. 

The R{\'{e}}nyi entropy is useful for measuring the information acquisition  cost since the parameter $\beta$ allows one to adjust the sensitivity of the cost to the shape of $\mu(x)$ and $\alpha_k(s|x)$. Indeed, R{\'{e}}nyi entropy of order $\beta$  includes the Hartley entropy, Shannon entropy, collision entropy and min entropy as special cases. In terms of (\ref{eqn:utilitymaximization}), the R{\'e}nyi information cost constrained decision problem is
\begin{align}
&p_k^*(x,a) \in\argmax_{p(x,a)} \Big\{\sum_{a\in\mathcal{A}_k}\sum_{x\in\mathcal{X}}p(x,a)u(x,a)\Big\} \nonumber\\
&\quad\text{s.t.}\quad \mu(x) = \sum_{a\in\mathcal{A}_k}p(x,a) \quad \forall x\in\mathcal{X} \nonumber\\
  &\phantom{\quad\text{s.t.}\quad} I_\beta(\mu,\alpha_k) \leq \kappa_\text{max},
\quad p(x,a) \geq 0 \quad \forall x\in\mathcal{X}, a\in\mathcal{A}_k.
\label{eqn:shannoninformationmax}
\end{align}
In (\ref{eqn:shannoninformationmax}), $\kappa_\text{max}$ represents the maximum ``effort'' the agent is willing to invest to estimate the state $x\in\mathcal{X}$ prior to taking the action $a\in\mathcal{A}_k$ in decision problem $k\in\{1,\dots,K\}$.

Given that the objective function is linear and the constraint set is convex in (\ref{eqn:shannoninformationmax}) for $\beta\in[0,1]$, necessary and sufficient conditions for the agent to satisfy rational inattention with the R{\'{e}}nyi information cost constraint can be constructed using the Karush-Kuhn-Tucker (KKT) conditions. Formally:
\begin{theorem}
A rationally inattentive agent with utility function $u(x,a)$, observed joint-distribution $p(x,a)$, and $\beta\in(0,1)$ satisfies R{\'{e}}nyi mutual information cost (\ref{eqn:renyimutualinformation}) if and only if there exists constants $\lambda_1 > 0$ and $\lambda_2$ that satisfy the linear constraints
\begin{align}
&u(x,a) = \frac{\lambda_1}{\beta-1}\eta^{\beta-1}(x,a)E[\eta^{\beta-1}(x,a)]-\lambda_2 \nonumber\\
&\frac{1}{\beta-1}\operatorname{ln}\left(\E[\eta^{\beta-1}(x,a)]\right) = \kappa_\text{max}, \qquad
\eta(x,a) = \frac{p(x|a)}{p(x)}
\end{align}
for all $x\in\mathcal{X}$, $a\in\mathcal{A}$ where $\E[\cdot]$ is the expected value taken over the joint-distribution $p(x,a)$. \qed
\label{thrm:reynicost}
\end{theorem}
In Theorem~\ref{thrm:reynicost},  $\lambda_1,\lambda_2$ are KKT multipliers of the R{\'{e}}nyi cost information constraint and  equality constraint in (\ref{eqn:shannoninformationmax}). Combining the linear equality constraints in Theorem~\ref{thrm:reynicost} with the mixed integer linear program (\ref{eqn:MILPagentutility}), yields a test for the R{\'{e}}nyi information cost constraint and  provides estimates of the associated  utility function of the agent. Thus we have constructed a preference based inverse reinforcement learning algorithm for the utility and information acquisition  cost of a Bayesian agent.

\section{Finite Sample Performance Analysis of the Agent's Action-Selection Policy}
\label{sec:Risk-AwareUtilityMaximizationbyOff-PolicyEstimation}
Thus  far we have constructed estimates for an agent's utility function and information acquisition  cost by observing the agents behavior.
Indeed, the maximum likelihood estimate of the agent's action-selection policy is $\hat{\pi}(a|x,f)$ (\ref{eqn:policyprior}).
An important  question related to performance analysis of these estimators  is: How far is the net utility obtained using this estimated policy (based on a finite dataset) compared
to the actual net utility $V(\boldsymbol{\pi}(a|x,f))$ (\ref{eqn:totalexpectedutility}) which uses the true policy $\boldsymbol{\pi}(a|x,f)$?

Using an extension of the empirical Bernstein inequality to the space of continuous function classes
\begin{align}
&\mathcal{F}_\Pi=\{f_{\pi,k}: \mathcal{X}\times\mathcal{A}_k\times{N}\rightarrow [0,1]\}, \; f_{\pi,k} = M\frac{\pi_k(a|x,f)}{\hat{\pi}_k(a|x,f)}u(x,a,f)=M\bar{u}(\pi_k(a|x,f))
\label{eqn:functionclass}
\end{align}
we can  construct a finite sample bound between the observed net utility $V(\hat{\boldsymbol{\pi}}(a|x,f))$ and an estimate of the net utility $V(\boldsymbol{\pi}(a|x,f))$ for the unobserved policy  $\boldsymbol{\pi}(a|x,f)$. In (\ref{eqn:functionclass}), $M$ is a normalization constant which ensures $f_{\pi,k}\in[0,1]$, $\hat{\pi}_k(a|x,f)$ is the observed policy (\ref{eqn:policyprior}), and $\pi_k(a|x,f)$ is an unobserved policy. By bounding the function class (\ref{eqn:functionclass}) using the uniform covering number and employeeing the double-sampling method~\cite{AB09}, Theorem~\ref{thrm:policyaccuracy} results. 
\begin{theorem}
Let $\bar{u}(\pi_k)$ be a random variable with $T_k$ i.i.d. samples in $\mathcal{D}$. Then with probability $1-\gamma$ the random vector $(a_t,x_t)\sim\pi_k$, for a stochastic hypothesis class $\pi_k\in\Pi$, $T_k \geq 16$, and $\lambda=\sqrt{18\operatorname{ln}(10\mathcal{N}_\infty\{1/T_k,\mathcal{F}_\Pi, 2T_k\}/\gamma)}$, satisfies
\begin{equation}
V(\pi_k) \leq \hat{V}(\pi_k)+\lambda\sqrt{\frac{\operatorname{Var}[\bar{u}(\pi_k)]}{T_k}}+\frac{15\lambda^2}{18M(T_k-1)}
\end{equation}
where $\mathcal{N}_\infty\{1/T_k,\mathcal{F}_\Pi, 2T_k\}/\gamma)$ is the uniform covering number. $\qed$
\label{thrm:policyaccuracy}
\end{theorem}
 Theorem~\ref{thrm:policyaccuracy}  provides a probabilistic bound between the estimated net utility $\hat{V}(\pi_k)$ and actual net utility $V(\pi_k)$ that only depends on the dataset $\mathcal{D}$ and the coefficient $\lambda$. Therefore, for constructing the true policy $\boldsymbol{\pi}(a|x,f)$, one would maximize the net utility $\hat{V}(\pi_k)$ while minimizing the variance term with a coefficient $\bar{\lambda} \geq 0$. Note that in Theorem~\ref{thrm:policyaccuracy} $\lambda$ encodes the entropy of the function class $\mathcal{F}_\Pi$,  which is dependent on the number of samples $T_k$, uniform covering number $\mathcal{N}_\infty\{\cdot\}$, and $\gamma$ which is a measure of the confidence of the estimate. For the function class (\ref{eqn:functionclass}), $\mathcal{N}_\infty\{\cdot\}$ is polynomial in the sample size $T_k$~\cite{MP09,VC15,Sau72}--this ensures as the sample size increases that $\hat{V}(\pi_k)\rightarrow V(\pi_k)$.

Using the insights from Theorem~\ref{thrm:policyaccuracy}, the mixed integer-linear program 
\begin{align}
&\boldsymbol{\pi}(a|x,f) \in \operatorname*{arg\, max}_{\pi_k\in\Pi}\left\{\sum_{k=1}^KV(\boldsymbol{\pi}_k(a|x,f))-\bar{\lambda}_k\sqrt{\frac{\operatorname{Var}[\bar{u}(\pi_k(a|x,f))]}{T_k}}\right\} \nonumber\\
&\text{ s.t. } \quad \sum_{a\in\mathcal{A}_k}\pi_k(a|x,f) = 1, \quad \pi_k(a|x,f) \geq 0 \nonumber\\
&  \mathcal{L}(u(a,x,f), \pi_k(a|x,f))  \quad \forall x\in\mathcal{X}, \forall a\in\mathcal{A}_k, \forall k\in\{1,\dots,K\}, \forall f\in\{1,\dots,N\}.
\label{eqn:offpolicymaximization}
\end{align}
can be used to construct the optimal policy $pi_k(a|x,f)$ that maximizes the net utility $V(\boldsymbol{\pi}(a|x,f))$ while ensuring the policy is consistent with rational inattention. The regularization term $\bar{\lambda}_k$ in (\ref{eqn:offpolicymaximization}) balances the maximization of the net utility $V(\boldsymbol{\pi}(a|x,f))$ while accounting for the finite-sample variance associated with estimating $V(\boldsymbol{\pi}(a|x,f))$ for policies $\boldsymbol{\pi}(a|x,f)$ that are different from $\hat{\boldsymbol{\pi}}(a|x,f)$. The lower the  value of $\bar{\lambda}_k$, the more risk-seeking the generated optimal policy. 

\section{Rational Inattention \&  Utility Maximization in YouTube Social Network}
\label{sec:RationalInattentionandUtilityMaximizationintheYouTubeSocialNetwork}

Constructing utility based preference models for how users  interact  and consume   content in online social media platforms is important in social network analysis \cite{Kha17,HG17}. YouTube is an interesting example of a social network since the interaction between users includes video content. Users interact on  YouTube channels  by  posting comments and rating videos. Extensive empirical  studies~\cite{Kha17,HG17,ABCH15,HNK15,HAK17,AK17} show  that comments and ratings from users are  influenced by the thumbnail, title, category, and perceived popularity of each video. Here we consider a massive YouTube dataset comprising 6 million videos across 25,000 channels and over a millions users from April 2007 to May 2015. As is typical in behavioral economics~\cite{WK17}, by user behavior, we mean the average commenting behavior per YouTube channel, averaged over all the channels. 

\begin{wrapfigure}{L}{0.5\textwidth}
  \centering
    \includegraphics[width=0.31\textwidth]{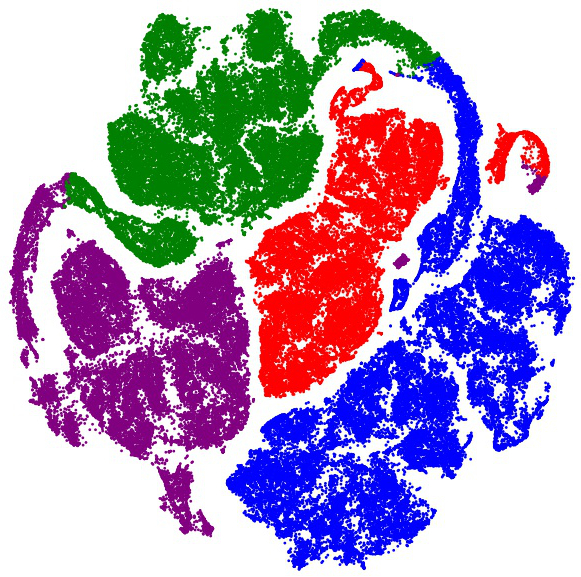}
    \vspace{0pt}
    \caption{t-Distributed Stochastic Neighbour visualization of the latent space representation of title and thumbnail constructed using Algorithm~\ref{alg:framingassociation} for the YouTube videos contained in the dataset $\mathcal{D}$. The association of each video to the four preference invariant frames is illustrated by four colors (red, blue, green, purple).}
    \label{fig:tsneunqiueframes}
\end{wrapfigure}

First, we constructed ordinal preference invariant frames using deep embedded clustering  Algorithm~\ref{alg:framingassociation}. Recall that Algorithm~\ref{alg:framingassociation}  maps the high dimensional title and thumbnail space to one of $N$ unique frames. Here we chose $N=4$ and the embedding space to have dimension $200$. The shape of the resulting embedding space is displayed in Fig.~\ref{fig:tsneunqiueframes}.
Selecting $N=4$ ensures  each video is sufficiently isolated to a particular frame; less than
3\% of videos are classified ambiguously  in terms of frames.

\begin{figure}[h]
  \centering
  \vspace{-9pt}
    \includegraphics[width=0.722\textwidth]{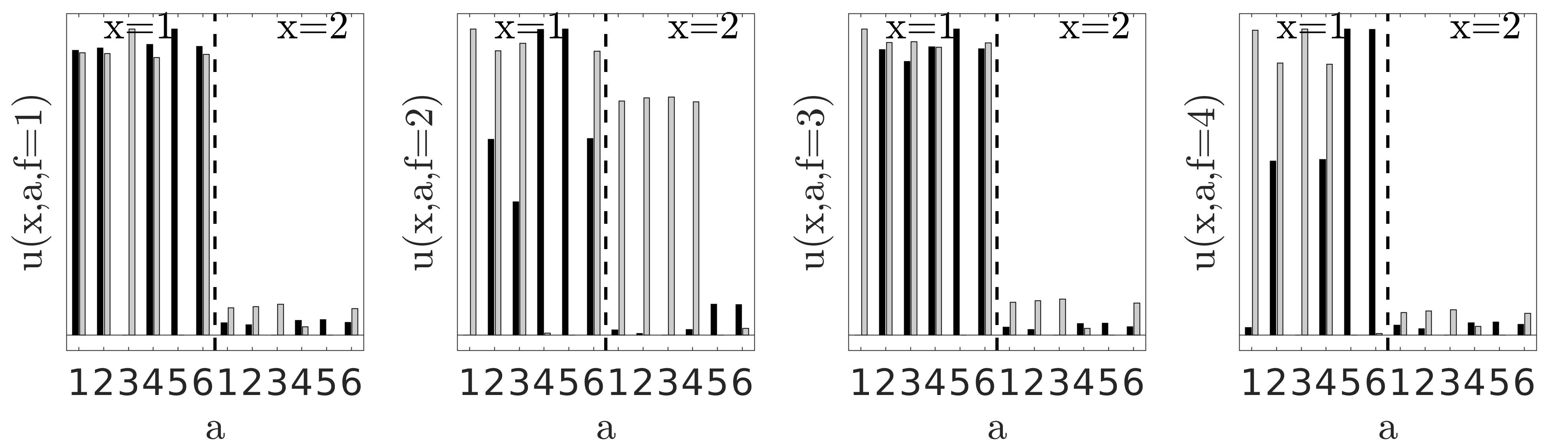}
    \vspace{-10pt}
    \caption{Utility function $u(x,a,f)$ of the rationally inattentive agents with a general information cost structure for each of the four unique frames in Fig.~\ref{fig:tsneunqiueframes}. Though the commenting preferences depend on the frame, the agent prefers to comment on videos with higher perceived popularity. The utility is constructed be evaluating the mixed-integer linear program (\ref{eqn:MILPagentutility}) with the YouTube dataset $\mathcal{D}$. $x$ represents the state, $a$ the possible actions, $f$ the frame, and the decision-problem $k$ indicates the most popular category (black bars) and the other categories (gray bars); see Appendix.}
\label{fig:YouTubeutilitygeneralcost}
\end{figure}

Next, for each of the preference invariant frames in Fig.~\ref{fig:tsneunqiueframes}, we apply the rational inattention test to determine if users are rationally inattentive. We find that the commenting behaviour of users in  YouTube is consistent with rational inattention for a general cost constraint. The ordinal utility of the users in each unique frame is provided in Fig.~\ref{fig:YouTubeutilitygeneralcost}. As expected, the commenting behaviour of the users is different between each frame. Additionally, the users prefer to comment on videos that are expected to have a higher popularity compared with videos with lower popularity. If we impose the R{\'{e}}nyi information cost constraint, we find that only the commenting behaviour in frame $f=4$ is rationally inattentive. The associated utility however provides no clear preference ordering between the popularity of the video and the associated commenting behaviour. This suggests that users are rationally inattentive with respect to a general information cost constraint.

{\em Discussion}. From a behavioral economics point of view, the above results yield useful insight into user behavior in online social multimedia.
Based on extensive analysis of the YouTube dataset, our main conclusions are that users commenting behavior (number of comments and comment sentiment) is i) consistent with rational inattention, ii) depends on the framing information available iii) users  prefer to comment on videos that are perceived to be popular, iv) the category of the video influences the commenting behavior; see Supporting Material. That deep clustering adequately captures  framing information, and that  a preference based utility with attention costs  rationalizes the YouTube dataset is remarkable. We speculate that this approach can be used to predict popularity
of YouTube channels.

There is also considerable scope to generalize the utility function estimation
described in this paper to  stopping time problems involving partially observed Markov decision processes \cite{Kri13,Kri16}.

\newpage

\appendix

\section{Appendix. Denoising Autoencoder Architecture for YouTube Title and Thumbnail}

A detailed description of the steps in the deep embedding method for constructing the preference invariant frames is provided in Algorithm 1, reproduced here in greater detail then in the main paper. The denoising autoencoder is comprised of stacked long short term memory (LSTM) and convolutional neural network (CNN) which are detailed in Sec.\ref{subsec:textprocessingYouTubetitle} and Sec.\ref{subsec:imageprocessingYouTubethumbnail}. To ensure the denoising autoencoder is robust to variations in the title and thumbnail input (e.g. good generalization performance), we introduce noise into the input training data. Possible methods to introduce noise into the network include using drop-out~\cite{SHKSS14} and drop-path~\cite{HSLSW16} methods. Here we apply Gaussian noise to the input images and numeric representation of the words, and additionally include drop-out layers in the LSTM and CNN networks. 

\begin{algorithm} 
\caption{Deep Embedded Clustering for Framing Association} 
\label{alg:framingassociation} 
\begin{algorithmic} 
    \REQUIRE Set of framing information $\{f_t\}_{t=1}^T$, number of unique frames $N$, stopping threshold $\delta\in(0,1)$, confidence threshold $\delta_c\in(0,1)$, and updating interval $\zeta$. 
    \HEADER{Pretrain}
    	\STATE Pretrain the denoising autoencoder without any frame association.
    \ENDHEADER
    \HEADER{Initialize}
    	\STATE Initialize the $N$ cluster centers $\Psi^o$ using k-means clustering in the latent space and set $\varepsilon=0$.
    \ENDHEADER
        \HEADER{Deep Clustering}
    	\STATE Train the deep clustering autoencoder and frame association layers. 
    	\STATE $i = 0$
    	\WHILE{$\sum_{t}n^o_t \neq n^i_t \geq T\delta$}
    		\STATE \IF{$i\%\zeta == 0$}
    			\STATE Compute all latent points $\{z_t=r(w(f_t))\}_{t=1}^T$
    			\STATE Compute $P$ using (9)
    			\STATE Set $n^o=n^i$
    			\STATE Compute new cluster labels $n^i_t = \operatorname*{arg\,max}_{n\in\{1,\dots,N\}}\{q_{in}\}$.
    		\ELSE
    			\STATE Select mini-batch sample from $\{f_t\}_{t=1}^T$ and update the weights of the autoencoder and frame association layers to minimize the loss (7). 
    		\ENDIF
    		\STATE $i=i+1$
    	\ENDWHILE
    \ENDHEADER
\RETURN Invariant frames $n_t \quad\forall t\in\{1,\dots,T\}$ such that $\operatorname{max}_n\{q_{tn}\} > \delta_c$. 
\end{algorithmic}
\end{algorithm}

\subsection{Text Processing of the YouTube Title}
\label{subsec:textprocessingYouTubetitle}
The design of autoencoders for text data is challenging as a result of the power-law distribution of words and the long-range dependencies (grammars) between words. To address these challenges, we use previously constructed word embeddings to convert the words into a numeric vector. We then  employ a LSTM networks for the encoder and decoder blocks of the autoencoder which focus on text processing. The combination of using word embeddings and LSTMs allows the network to utilize prior knowledge of similar words while simultaneously learning how to cluster similar sentences into a unique frame. 

Prior to transforming the words into their numeric embedding, we apply a lemmatization transformation.  Lemmatization reduces the number of variations of words necessary to consider as it groups all the inflected forms a word into a single base representation. For example, the verb ``to walk'' may appear as ``walk'', ``walked'', ``walks'', ``walking'' which are all converted to ``walk'' via the lemmatization transformation. To perform the lemmatization transformation we use the WordNet lemmatizer~\footnote{https://wordnet.princeton.edu/wordnet/}. The WordNet lemmatizer is comprised of two resources, a set of rules which identify the inflectional endings that can be detached from individual words, and a list of exceptions for irregular word forms. WordNet first checks the exceptions, then remove any inflectional endings from the words. Having performed the lemmatization operation, we now construct numeric vector representations of the words. A popular method to perform this task is to use distributed representations of words (e.g. word embeddings). The distributed representation of words in a vector space are designed such that words with similar semantic meaning have similar latent space representations. Equivalently, words with similar meaning will cluster tgeother in the word embedding space. Two popular word embeddings are the Word2Vec~\cite{MSCCD13} and Glove~\cite{PSM14} models. For the clustering algorithm we use the Glove embedding that was constructed using over 2 billion tweets and is comprised of over 1.2 million words. The possible dimension of the word embedding space is 25, 50, 100, or 200. Here we use a word embedding dimension of 25. 

Given the word embeddings of the sentence $w(f)$, we use an LSTM encoder-decoder framework to learn latent space representations of the titles~\cite{Gol16,SVL14,GBC16,Ger17}. To construct the latent space representation of the sentences, we utilize a stacked LSTM architecture. Note that stacked LSTMs are able to capture grammatical information in the title at different scales. It was illustrated in~\cite{Gol16,SVL14} that stacked LSTMs tend to have superior predictive performance compared to single layer LSTMs for natural language processing tasks.

\subsection{Image Processing of the YouTube Thumbnail}
\label{subsec:imageprocessingYouTubethumbnail}
In the denoising autoencoder, image processing is performed using a VGG based architecture. Given the latent space representation $z_t$ from the encoder, the image decoder is used to reconstruct the original input image. To perform this task requires the use of deconvolution and upsampling layers. However, deconvolution layers are not used in CNN autoencoders. Instead a mixture of convolutional and upsampling layers are employed. In the most extreme case, a single upsampling layer can be used to directly reconstruct the images from the latent space as illustrated in~\cite{LSD15}. A commonly used method is to construct multiple transposed convolution (also known as fractionally strided convolutions) layers in combination with upsampling layers. Using the transposed convolution layers instead of the standard convolution layers ensures that ``checkerboard'' artifacts are removed from the decoded image~\cite{ODO16}.

\section{Constraint Set $\mathcal{L}(u(x,a,f)$ for Rational Inattention (Theorem~1) and Recovery of Utility and Information Cost}
To construct the utility function $u(x,a,f)$ of the agent for the observed stochastic dataset $\mathcal{D}$ (4) requires that the utility satisfies the inequalities (10) for Bayesian utility maximization, and (11) for attention function maximization. The utility function $u(x,a,f)$ of a rationally inattentive agent must satisfy the following mixed-integer linear constraints: 
\begin{align}
&\sum_{x\in\mathcal{X}}p_k(x|a,f)[u(x,a,f)-u(x,b,f)] \geq 0 \label{eqn:MILPagentutility}\\
&\sum_{k=1}^K\left(\sum_{a\in\mathcal{A}_k}p_k(a,f)m_{k}(a,f)-\sum_{a\in\mathcal{A}_{k+1}}p_{k+1}(a,f)n_{k+1}(a,f)\right) \geq 0 \nonumber\\
&m_k(a,b,f) = \sum_{x\in\mathcal{X}}p_k(a|x,f)u(x,b,f) \nonumber\\
&m_k(a,f) \geq m_k(a,b,f) \quad \forall a,b\in\mathcal{A}_k \nonumber\\
&m_k(a,f) \leq m_k(a,b,f)_M(1-\delta_{b,f}), \quad \sum_{b\in\mathcal{A}_k}\delta_{b,f} = 1 \nonumber\\
&n_{k+1}(a,f) \geq m_{k+1}(a,b,f) \quad \forall a\in\mathcal{A}_{k+1} \quad \forall b\in\mathcal{A}_k \nonumber\\
&n_{k+1}(a,f) \leq m_{k+1}(a,b,f)_M(1-\zeta_{b,f}), \quad \sum_{b\in\mathcal{A}_k}\zeta_{b,f} = 1 \nonumber\\
&\quad u(x,a)\in[0,1], \quad \delta_{b,k}, \zeta_{b,k} \in\{0,1\} \nonumber\\
&\forall a,b\in\mathcal{A}_k, \quad c\in\mathcal{A}_{k+1}, \quad \forall k\in\{1,2,\dots,K\} \nonumber
\end{align}
with $\mathcal{A}_{K+1}=\mathcal{A}_1$ and $M$ a large constant. To determine if a $u(x,a,f)$ exists for the constraint set can be evaluated using a variety of numerical methods including branch-and-bound, cutting planes, branch-and-cut, and branch-and-price~\cite{GG11}.


Given the utility function $u(x,a,f)$ from the solution of (12), and the inequality relation (13), an ordinal estimate of the associated cost of information $C(\mu,\alpha_k)$ of each attention strategy $\alpha_k$ can be constructed. Specifically, the ordinal cost of information $C(\mu,\alpha_k)$ can be computed by solving the following linear program: 
\begin{align}
&G_{k,k}-G_{w,k} \geq C(\mu,\alpha_k)-C(\mu,\alpha_w) \nonumber\\
&C(\mu,\alpha_k) \geq 0 \forall w,k\in\{1,\dots,K\}.
\label{eqn:ordinalcost}
\end{align}
Recall that if a solution to (12) exists, then a solution to (\ref{eqn:ordinalcost}) is guaranteed to exist from Theorem~1 and (3). Notice that if the cost of a particular attention strategy is zero, then absolute bounds can be placed on the information cost of each attention strategy. For example if $C(\mu,\alpha_w)=0$, then the cost $C(\mu,\alpha_k) \in [G_{k,w}-G_{w,w}, G_{k,k}-G_{w,k}]$. The estimated cost function satisfies weak monotonicity in information--that is, if the attention function provides more information then it will have a higher information cost. However, it may be the case that the actual cost of information used by the agent does not satisfy this condition. In fact, only requiring rational inattention with no further restrictions on information cost does not impose any testable conditions for information monotonicity.

\section{Estimating the Agent's Attention Function and Choice Function}
If the dataset $\mathcal{D}$ satisfies rational inattention, it is also possible to estimate the agent's attention function $\alpha_k(s|x)$ and choice function $\eta_k(a|s)$.

To construct the agent's attention function $\alpha_k(s|x)$ and choice function $\eta_k(a|s)$ requires the posterior distribution $p_k(x|a)$. First, consider the signal set $\mathcal{S}(\alpha_k)$ of all observed posterior state distributions of the agent for attention function $\alpha_k(s|x)$ using
\begin{equation}
\mathcal{S}(\alpha_k) = \{p_k(x|a): a\in\mathcal{A}_k\}, \quad \label{eqn:revealedposterior}
p_k(x|a) = \frac{\mu(x)\pi_k(a|x)}{\sum_{y\in\mathcal{X}}\mu(y)\pi_k(a|y)}.
\end{equation}
Each posterior distribution $p_k(x|a)$ is associated with a single signal $s\in\mathcal{S}(\alpha_k)$. The posterior distribution $p_k(x|a)$ in (\ref{eqn:revealedposterior}) is equal to the true posterior distribution $p_k(x|s)$ in (1) only if the choice function $\eta_k(a|s)$ produces a single action $a\in\mathcal{A}_k$ for each $s\in\mathcal{S}(\alpha_k)$ with probability one. Otherwise the posterior distribution $p_k(x|a)$ is given by the weighted sum
\begin{equation}
p_k(x|a) = \frac{\sum_{s\in\mathcal{S}(\alpha_k)}\eta_k(a|s)p_k(x|s)p_k(s)}{\sum_{x\in\mathcal{X}}\sum_{s\in\mathcal{S}(\alpha_k)}\eta_k(a|s)p_k(x|s)p_k(s)}.
\end{equation}
Note that without explicit knowledge of the choice and attention functions of the agent, the stochastic choice dataset can not be used to determine if $p_k(x|a)=p_k(x|s)$.  Having $p_k(x|a)=p_k(x|s)$ is not required to determine if the agent satisfies rational inattention. 

Given $p_k(x|a)$, for each signal $s\in\mathcal{S}(\alpha_k)$, the associated attention function is
\begin{equation}
\alpha_k(s|x) = \sum_{a\in\mathcal{A}_k}\eta_k(a|s)\alpha_k(s|x)
=\sum_{a\in\mathcal{A}_k}\pi_k(a|x)\mathbf{1}\{p_k(x|a) = s\}
\label{eqn:attentionfunction}
\end{equation}
where the second equality results from using the data matching condition in Theorem~1. Note that (\ref{eqn:attentionfunction}) is only equal to the agent's attention function $\rho_k(r|x)$ if the observed and true posterior distributions are equal. If $\rho_k(r|x)$ is the true attention function then
\begin{align}
\alpha_k(s|x) &= \sum_{r\in\mathcal{S}(\rho_k)}\sum_{a\in\mathcal{A}_k}\eta_k(a|r)\rho_k(r|x)\mathbf{1}\{p_k(x|a) = s\}.
\end{align}
It must be the case that the observed attention strategy $\alpha_k(s|x)$ is weakly less informative than the true attention strategy $\rho_k(r|x)$. Equivalently, the observed attention strategy is a noisy version of the true attention strategy. Theorem~1 however does not require we know the true attention strategy $\rho_k(r|x)$ of the agent to test if the agent's behavior satisfies rational inattention. 

The observed choice function of the agent is given by
\begin{equation}
\eta_k(a|s) = \frac{\sum_{x\in\mathcal{X}}\mu(x)\pi_k(a|x)}{\sum_{b\in\mathcal{A}_k}\sum_{x\in\mathcal{X}}\mu(x)\pi_k(b|x)\mathbf{1}\{p_k(x|b) = s\}}
\label{eqn:choicefunction}
\end{equation}
which is merely the ratio of the number of times action $a\in\mathcal{A}_k$ was selected over all other possible actions $b\in\mathcal{A}_k$ for the prior distribution $s\in\mathcal{S}(\alpha_k)$. The observed choice function provides no information on the true choice function over the posterior distributions $r\in\Gamma(\rho_k)$ that result from the true attention function unless the actual and observed posterior distributions are equal. Note however that the observed attention function $\alpha_k(s|x)$ (\ref{eqn:attentionfunction}) and choice function $\eta_k(a|s)$ (\ref{eqn:choicefunction}) are consistent with the agent's observed action-selection policy $\pi_k(a|x)$ as required in the data matching requirement of Theorem~1.

\section{YouTube Dataset and Definition of the Frames, Context, Action, and Decision-Problem}

To construct $\mathcal{D}$, we use the real-world YouTube dataset comprising 6 million videos across 25,000 channels from April 2007 to May 2015. The YouTube data contains the view counts, comment counts, likes, dislikes, thumbnail, title, and category of each video. The frame instance $f_t$ of each video is comprised of the video's thumbnail and title. Specifically, we use a $40\times80$ pixel color image to represent the thumbnail (which is a resized version of the native $246\times138$ pixel thumbnails used in YouTube). For the title, we only include the first 8 words of the title in the framing instance $f_t$ (over 90\% of the videos have a title of length 8 words or less). The top category of videos in the YouTube dataset is ``Gaming'' which comprises 44\% of all the videos. Two decision-problems are considered in the dataset. The first is $k=1$ which is associated with all videos that have category ``Gaming'', while decision-problem $k=2$ results for videos that are not associated with the ``Gaming'' category. The state $x_t$ of each video is associated with the viewcount of the video 14 days after the video was published. Specifically, state $x=1$ is high viewcount where the viewcount is above 10,000 views, while $x=2$ results otherwise. The associated action $a_t$ is related to the commenting behavior of the agents, which is computed using the comment counts, like count, and dislike count 2 days after the video is published. The possible actions $a=1$ is low comment count with negative sentiment, $a=2$ is low comment count with neutral sentiment, $a=3$ is low comment count with positive sentiment, $a=4$ is high comment count with negative sentiment, $a=5$ is high comment count with neutral sentiment, and $a=6$ is high comment count with positive sentiment. Here negative sentiment results if the difference in like count and dislike count is below -25, neutral sentiment if the difference is between -25, 25, and has positive sentiment if the difference is above 25. A low comment count is considered if there are less then 100 comments, and high otherwise.

\bibliographystyle{plain}

\bibliography{inattention}

\end{document}